\documentclass[conference]{IEEEtran}
\ifCLASSINFOpdf
\else
\fi
\usepackage{graphicx}

\begin{document}
\title{Supervised Classification of RADARSAT-2  Polarimetric Data for Different Land Features}

\author{
\IEEEauthorblockA{{\large Abhishek Maity}}
}


%Telephone: (800) 555--1212, Fax: (888) 555--1212}


\maketitle

\begin{abstract}
The pixel percentage belonging to the user defined area that are assigned to cluster in a confusion matrix for RADARSAT-2 over Vancouver area has been analysed for classification. In this study, supervised Wishart and Support Vector Machine (SVM) classifiers over RADARSAT-2 (RS2) fine quad-pol mode Single Look Complex (SLC) product data is computed and compared. In comparison with conventional
single channel  or dual channel polarization, RADARSAT-2 is fully polarimetric, making it to offer better land feature contrast for classification operation.\\
\end{abstract}


% \fi
\IEEEpeerreviewmaketitle

\section{Introduction}
SAR data analysis for a range of applications from compact and fully Polarimetric SAR like RADARSAT-2 are becoming popular everyday on the fact that they offer features like higher resolution imaging, wide swath, reduced PRFs. Data from these family of SAR are very useful in several applications involving terrain to oceans which are clearly depicted in\cite{charbonneau2010cjrs}. The polarimetric data analysis from Convair-580 and RADARSAT-2 have resulted many successful studies in fields ranging from ship-detection\cite{touzi2013apsar} , land-use pattern, crop classification.\par With the launch of RADARSAT-2 on December 14, 2007, it became possible to have a SAR system having modes of multiple polarization including full polarimetry and resolution upto 1 metre in spotlight mode. The satellite carries a C-band SAR. In this paper, confusion matrix analysis of RADARSAT-2 data has been examined for various feature classification.\par Many studies have been undertaken for classification using RADARSAT-2 till date. They include classification of terrain classes using Random Forest\cite{du2015isprs}, ship detection\cite{touzi2013apsar}\cite{touzi2015tgrs}, oil slick characterization\cite{staples2014iosc}, crop monitoring of rice in China\cite{wu2011grsl} and identification of potato and rice fields using RADARSAT-1 in India\cite{panigrahy1999isprs}.Even work like ice-monitoring\cite{scheuchl2004cjrs} and mapping of seasonal floods in wetland forests of Brazil has yielded promising results\cite{evans2010jstars}. Research on classification using multi-temporal data sets of RADARSAT-2 is also being carried out using different classification algorithms coupled with various functions.

\section{Vancouver study site: RS2 Data}
Here for the study, we utilize the data set consisting the Greater Vancouver area, Canada. The test site is very diverse in nature consisting a wide variety of features to classify. The area consists of  urban settlements including Richmond area, rotated urban areas west to New Westminster. Rugged mountains in northern Vancouver, rivers merging to the Strait of Georgia and crop-lands in the Fraser River Delta. \par RADARSAT-2 data has been acquired on May 2008 over Vancouver area in full polarimetric mode. Ground-truth parameters was also collected synchronous with the satellite pass. Near Range Incidence Angle is $ 34.49^{o} $ and Far Range Incidence Angle is $ 36.08^{o} $.The dataset of RADARSAT-2 was acquired in Fine Quad mode with Q15 beam. It has been captured in descending pass direction inferring the snap is recorded on the sunlit side as the orbit of the SAR system is sun-synchronous

\begin{figure}
	\centering
	\includegraphics[width=1.0\linewidth]{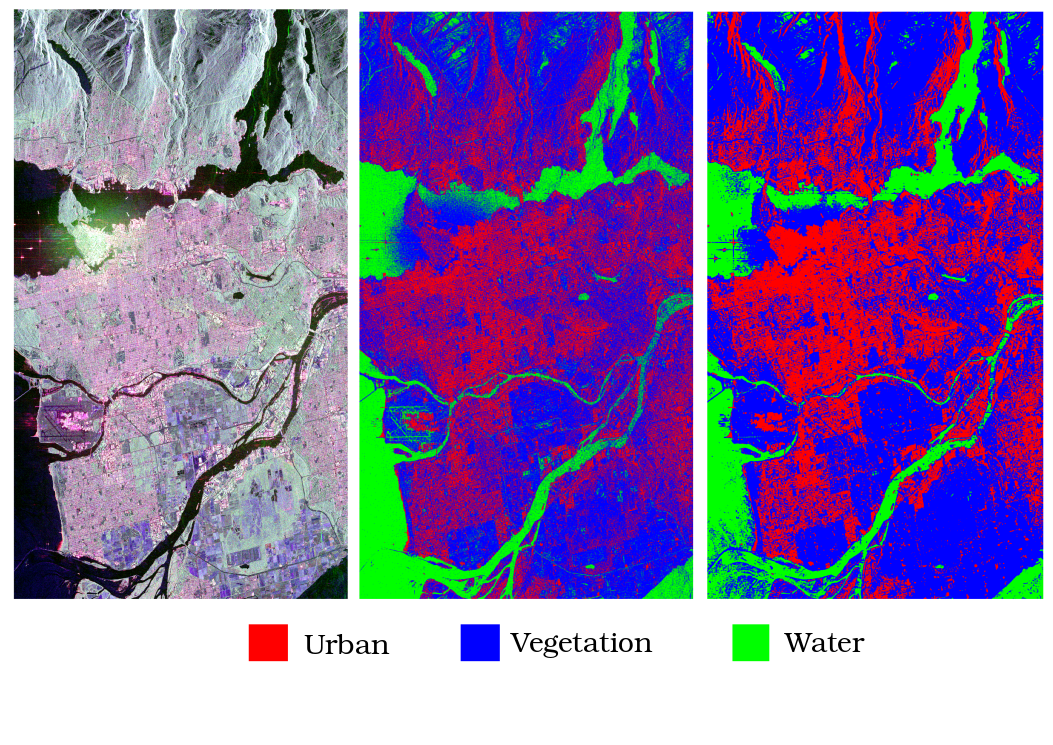}
	\caption{(a) PauliRGB and classification based on supervised (b) SVM (c) Wishart classifiers}
	\label{fig:Pauli_SVM_Wishart}
\end{figure}

\section{Methodology}
  
The coherency matrix \textbf{T} of $ 3 \times 3 $ is generated. Beside $T_{3}$ matrix, additional polarimetric information like $ H/A/\alpha $ coefficients for Target decomposition are computed for performing classification based on Wishart distribution \cite{lee1994ijrs}.

\par Speckle is a kind of noise that appears in data obtained through SAR systems. Speckle reduces with multi-looking images. Therefore, we have applied a $ 3 \times 3 $ look using Lee filter to remove noise keeping loss of information minimum.

The coherency matrix T is computed through a scattering
vector in the base of Pauli that demonstrate geometrical properties.The equation is given by and is in accordance to \cite{cloude1995tgrs}

\[ k_{p}=\frac{1}{\sqrt{2}}\left( \begin{array}{c}
S_{HH}+S_{VV} \\S_{HH}-S_{VV} \\2S_{HV}  \end{array} \right)\]

\[[T]=k_{p}.k_{p}^{*T}\]

After decomposition from \cite{cloude1995tgrs} the $ H $ entropy shows the wave polarization, where as $ A $ or Anisotropy is a difference between
the second and third eigenvalue especially significant for the range $  0.7 < entropy < 0.9 $. The $ \alpha $ parameter is an important component because it gives the wave reflection mechanisms over the considered
pixels. It characterizes the single bound, double bound and volume scattering. 

\subsection{Wishart Classification} 

The T matrix elements especially dedicated to SAR data involves the Wishart classification as because the presence of speckle noise in the data set account for the Wishart distribution. The polarimetric information for  mono-static case is define by the target vector $ h $

\[ h= \left( \begin{array}{c}
S_{HH} \\\sqrt{2}.S_{HV} \\S_{VV}  \end{array} \right)\]

In the multi-look data that is $ 3 \times 3 $ case we represent
the data by a polarimetric covariance matrix $ Z $

\[ Z=\frac{1}{n}.\sum\limits_{k=1}^{n}h_{k}h_{k}^{*T} \]

Where $ h_{k} $ is nothing but $ k $th sample of $ h $, the superscript * in the equation denote the complex conjugate where as the number of looks (samples) is given by $ n $. As per Wishart distribution the covariance matrix could be expressed as :

\[ p(Z)=\frac{n^{qn}|Z|^{n-q}exp^{-tr(n\sum^{-1}Z)}}{K(n,q)|\sum|^{n}} \]
with \[ K(n,q)=\pi^{q(q-1)/2}.\prod\limits_{i=1}^{q}\Gamma(n-i+1) \]

Where $ \Gamma() $ represents the gamma function and $ tr() $ is the trace of the given matrix. The $ q $ denotes the number of elements of the obtained target vector $ h $ (It is generally 3 for mono-static and 4 for the bi-static). Lastly, the $ n $ represent the number of looks. It is to be noted that the Wishart classification consist in a maximum likelihood classification based on a Wishart distribution.

\subsection{SVM classification}

The Support Vector Machine (SVM) are models of supervised learning that basically analyse data used for classification and regression analysis. The work here coincide with \cite{chang2011tist} and \cite{burges1998dmkd}.

\begin{figure}
	\centering
	\includegraphics[width=1.0\linewidth]{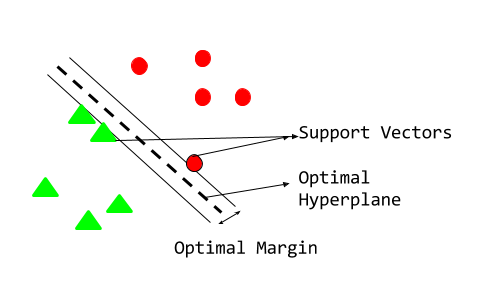}
	\caption{Linear SVM classifier}
	\label{fig:linear_SVM}
	\includegraphics[width=1.0\linewidth]{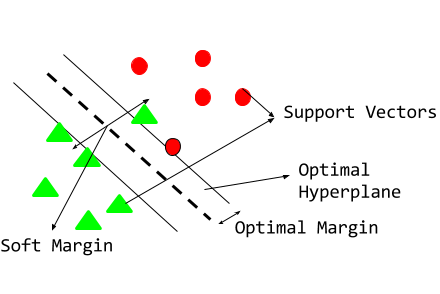}
	\caption{Non-linear SVM classifier}
	\label{fig:non-linear_SVM}
\end{figure}

\par \emph{Linear case:} With $ N $ training samples, the case of two
classes problem is considered. Every sample is described
by a Support Vector $ X_{i} $ consisting of different
"band" having $ n $ dimensions.  A sample is labelled as $ Y_{i} $. Here, we shall consider the first class label as -1 and other as +1. The SVM classifier consist in defining the function

\[ f(x) = sign(\langle \omega, X\rangle + b)  \]
that found the optimum separating hyperplane as presented in

\par The label from sample gives the sign of $ f(x) $. The target of the SVM is to maximize the margin between the support vector and the optimal hyperplane. Thus, we look for the $ min\frac{||\omega||}{2} $. For executing this, we tend to use the Lagrange multiplier

\[ f(x) = Sign(\sum\limits_{i=1}^{Ns}y_{i}.\alpha_{i} \langle x, x_{i}\rangle + b) \]
where Lagrange multiplier is $ \alpha_{i} $.

\par \emph{Nonlinear case:} In non-linear as the Fig.  the solution involves first to develop soft margin that is adapted to data containing noise. The next solution of SVM is to utilize a kernel. The kernel in this context is a function where the projection of the initial data is simulated in a space feature with greater dimension $ \phi:{R}^n\longrightarrow h $ . In this new space the information are considered as separable linearly. Thus, the dot product $ \langle x_{i}, x_{j}\rangle $ is replaced by

\[ K(x,x_{i}) =\langle \phi(x),\phi(x_{i})\rangle\]

The classification turns to be
\[ f(x) = Sign(\sum\limits_{i=1}^{Ns}y_{i}.\alpha_{i}.K(x,x_{i}) + b) \]
In general three types of kernels are used\\1. Polynomial kernel \[K(x,x_{i})= (\langle \phi(x),\phi(x_{i})\rangle +1)^{p} \] 2. Sigmoid kernel \[ K(x,x_{i})=\tanh (\langle \phi(x),\phi(x_{i})\rangle +1) \] 3. RBF kernel \[ K(x,x_{i})=exp^{-\frac{|x-x_{i}|^{2}}{{2\sigma^{2}}}} \] 
\par In accordance to the nature of this work, the RBF kernel is used as because it yields the best result.

\par For classification using supervised Wishart and SVM classifiers, $T_{3}$ matrix elements of SAR data is processed. For SVM, lib-SVM \cite{chang2011tist} is applied. The Radial Basis Function (RBF) kernel $ \gamma=1/\sigma $ is 0.444 and the cost is 100. In both the processes, we select training areas in accordance with the ground truth. Nine test areas representing the type of terrain cover present in the area were selected. The classes included urban, water and non-urban. Significant pixel density were selected for every class and the standard and mean deviation of the back-scattering were also calculated. Same training set is used for both the classifiers. Training cluster maps are generated for different classes for each classifier. Confusion matrix is calculated and generated for every class from SVM and Wishart polarimetric segmentation.

\section{Results and Discussions}

\begin{table}[!ht]
	\begin{center}
		\begin{tabular}{ l | c c r }
			\hline
			CLASS & Urban & Vegetation & Water \\
			\hline
			Urban & 87.78 & 0 & 12.22\\
			Vegetation & 0 & 99.95 & 0.05\\
			Water & 9.41 & 0.16 & 90.43\\
			\hline\\
		\end{tabular}
	\caption{Wishart Confusion matrix with overall classification accuracy (in \%)}

		\begin{tabular}{ l | c c r }
			\hline
			CLASS & Urban & Vegetation & Water \\	 	 
			\hline
			Urban & 72.53 & 0.25 & 27.22\\
			Vegetation & 0 & 97.70 & 2.30\\
			Water & 10.53 & 6.17 & 83.30\\
			\hline
		\end{tabular}
	\end{center}
	\caption{SVM Confusion matrix with overall classification accuracy (in \%)}
\end{table}
The rows represent the user defined clusters columns represent the segmented clusters. A number located at a position $ (I,J) $ represents the amount of pixels in percent belonging to
the user defined area $ I $ that were assigned to
cluster $ J $ during the supervised classification. The results through the confusion matrices shows that the performance by the Wishart is little better than the SVM. But generally SVM is the best from the other as we see from \cite{lardeux2007pollinsar}.This may be because of the training areas computed for classification.

\section{Conclusion and Future Work}
Fully Polarimetric data has significant contribution for urban and tropical vegetation cartography. For full polarimetric mode (swath 2x bigger), dual Polarimetry and particularly $ \pi/4 $ turns out to be a good compromise .Wishart can be a good potential for PolSAR data classification for very few cases.


\section*{Acknowledgment}
The author would like to thank Mr. Shaunak De for helping to understand on SAR and MDA corporation for providing RS2 sample data of Vancouver site.



%\bibliographystyle{IEEEtran}

\end{document}